\documentclass{article}

\usepackage{PRIMEarxiv}

\usepackage[utf8]{inputenc} 
\usepackage[T1]{fontenc}    
\usepackage{hyperref}       
\usepackage{url}            
\usepackage{booktabs}       
\usepackage{amsfonts}       
\usepackage{nicefrac}       
\usepackage{microtype}      
\usepackage{lipsum}
\usepackage{fancyhdr}       
\usepackage{graphicx}       
\graphicspath{{media/}}     

\title{Accelerating Local LLMs on Resource-Constrained Edge Devices via Distributed Prompt Caching
}

\author{
  Hiroki Matsutani, Naoki Matsuda, and Naoto Sugiura \\
  Dept. of ICS, Keio University \\
  3-14-1 Hiyoshi, Kohoku-ku, Yokohama, Japan\\
  \texttt{\{matutani,matsuda,nsugiura\}@arc.ics.keio.ac.jp} \\
}

\begin{document}
\maketitle

\begin{abstract}
Since local LLM inference on resource-constrained edge devices imposes a severe performance bottleneck, this paper proposes distributed prompt caching to enhance inference performance by cooperatively sharing intermediate processing states across multiple low-end edge devices.
To fully utilize prompt similarity, our distributed caching mechanism also supports partial matching.
As this approach introduces communication overhead associated with state sharing over a wireless network, we introduce a Bloom-filter-based data structure, referred to as a catalog, to determine whether a remote server possesses the desired internal states, thereby suppressing unnecessary communication.
Experiments using the Gemma-3 270M model and the MMLU dataset on the Raspberry Pi Zero 2W platform demonstrate that the proposed approach reduces TTFT (Time to First Token) and TTLT (Time to Last Token) by 93.12\% and 50.07\% on average, respectively.
\end{abstract}


\section{Introduction}\label{sec:intro}
A local LLM (Large Language Model) refers to the deployment of an LLM in a local environment, in contrast to cloud-based LLM services; typical examples include those running on on-premises servers and mobile devices.
Although local LLMs are often limited in size due to available compute resources, they offer distinct benefits.
First, as queries and prompts are processed in a private environment, security and privacy concerns can be effectively addressed; this is particularly appealing for applications dealing with privacy-sensitive data, such as personal assistants and surveillance cameras.
Second, as services do not rely on the cloud or Internet connectivity, response times are more predictable and typically offer low latency; this is well-suited for control applications in edge environments.

The expansion of local LLM applications to the edge necessitates deployment on resource-constrained platforms.
For instance, we have demonstrated local LLMs on Raspberry Pi Zero 2W, which is known as a \$15 computer, targeting embedded and wearable applications \footnote{\url{https://www.youtube.com/watch?v=PjXGlZcVzDA}}\footnote{Gemma-3 270M model is used for the demonstration. Although the model performance is limited, it is useful as a base model for specific applications through fine-tuning.}.
Accordingly, this paper focuses on this class of low-cost edge computers.
However, as demonstrated in Section \ref{sec:eval}, the performance is quite limited due to constrained compute resources, requiring tens of seconds for simple queries; this negates the benefits of local LLMs mentioned above.

\begin{figure}[t]
  \centering
  \includegraphics[height=56mm]{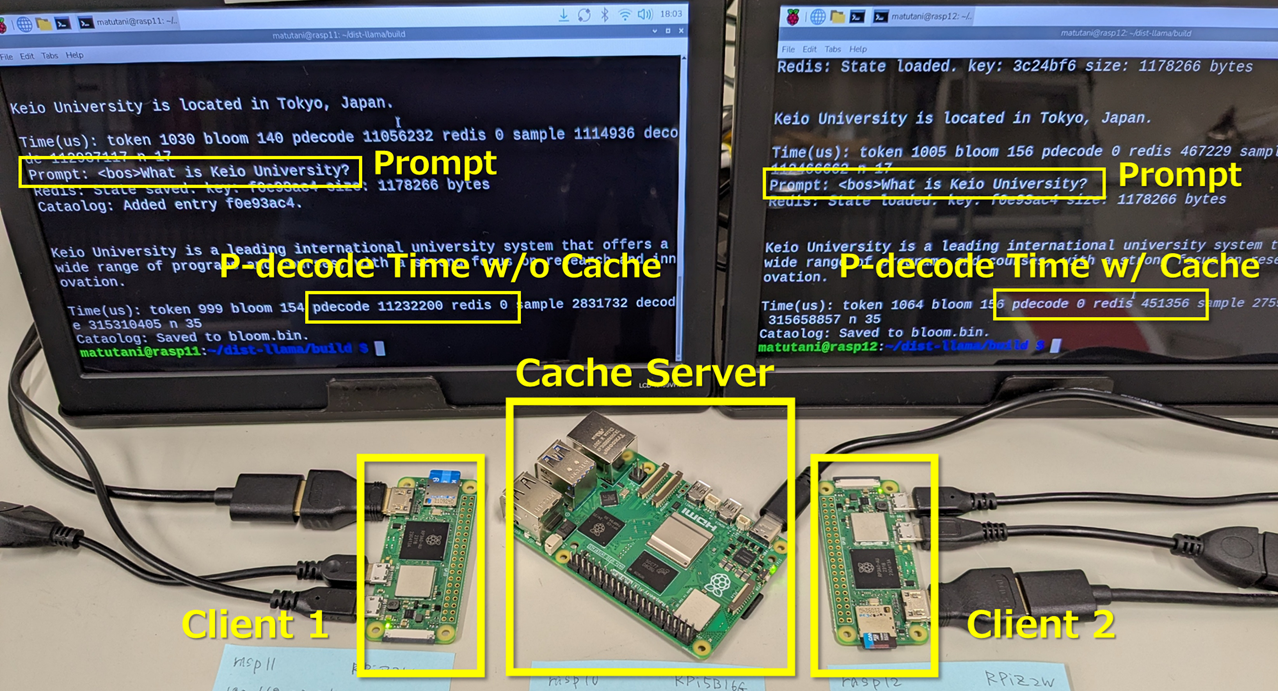}
  \caption{System overview. A local LLM on Client 1 (left) first processes a prompt and uploads its internal states to Cache Server (middle), accelerating a subsequent query on Client 2 (right).}
  \label{fig:system}
\end{figure}

To overcome this limitation, this paper proposes distributed prompt caching to enhance the performance of local LLMs, specifically Time to First Token (TTFT), on resource-constrained edge devices.
This approach enables the cooperative sharing of intermediate processing states across devices, leveraging prompt similarity.
As shown in Figure \ref{fig:system}, for instance, a local LLM on Client 1 (left) first processes a prompt and uploads its internal states to the Cache Server (middle), thereby accelerating a subsequent query on Client 2 (right).
The contributions of this work are summarized as follows:
\begin{itemize}
\item We identify the performance limitations of local LLMs on low-end edge devices using the MMLU dataset.
\item We extend the concept of prompt prefix caching to a multi-device environment to address these performance bottlenecks. 
\item We introduce a local {\it catalog} to effectively eliminate the communication overhead associated with state sharing over a wireless network.
\item We analyze the break-even point of the proposed approach on real edge hardware to evaluate its practical feasibility.
\end{itemize}

This paper is organized as follows.
Section \ref{sec:survey} briefly reviews caching techniques for LLM inference.
Sections \ref{sec:design} and \ref{sec:imple} present the design and implementation of the proposed distributed prompt caching mechanism.
Section \ref{sec:eval} discusses experimental results, and Section \ref{sec:conc} concludes the paper.

\section{Background and Related Work}\label{sec:survey}
A typical LLM architecture consists of dozens of Transformer blocks, each primarily comprising Multi-Head Attention (MHA) and Feed-Forward Network (FFN) layers \cite{Vaswani17}.
In particular, MHA is the core component that captures dependencies within a sequence of past tokens to enable next-token prediction; this process is known as autoregressive token generation.
In the autoregressive token generation process, the Key-Value (KV) cache \cite{Pope23} is a widely used technique for storing the K and V tensors of past tokens, thereby avoiding redundant computations when generating subsequent tokens.

While conventional KV caches are typically maintained within a single context, they can also be reused across multiple contexts; such a technique is known as prompt caching \cite{Gim24}.
This allows systems to bypass all or part of the prompt decoding process when input prompts share a substantial overlap, such as when they are generated from the same template.
To further increase cache reuse opportunities, research primarily follows two directions: prefix caching and semantic caching. 
In prefix caching, common prefixes are pre-defined by service providers in \cite{Kwon23}, whereas they are dynamically managed using a prefix-tree structure in \cite{Ye24}.
Similarly, the KV cache is managed as a radix-tree structure in the SGLang framework \cite{Zheng24}.
Semantic caching, on the other hand, retrieves reusable KV states based on the cosine similarity between the incoming and cached prompts \cite{Regmi24}.

In addition, various research efforts have explored enhancing KV cache efficiency.
For instance, the caches are compressed to accelerate LLM serving \cite{Liu24}.
To reuse KV caches across multi-turn conversations, the caches are saved in cost-effective memory and storage mediums \cite{Gao24}.

These approaches mentioned above are promising, particularly for LLM servers that process a high volume of queries from clients.
In the context of edge-based LLMs, this is especially relevant in scenarios where edge devices serve as clients querying remote servers; in such cases, LLM services rely heavily on server-side infrastructure.
In contrast, this paper focuses on local LLM inference on resource-constrained edge devices, where LLM tasks are executed on-device.
To this end, we propose a distributed prompt caching mechanism to enhance local inference efficiency by extending the concept of prompt caching to a cooperative framework across multiple low-end edge devices.

\section{Distributed Prompt Caching Mechanism}\label{sec:design}

\begin{figure}[t]
  \centering
  \includegraphics[height=62mm]{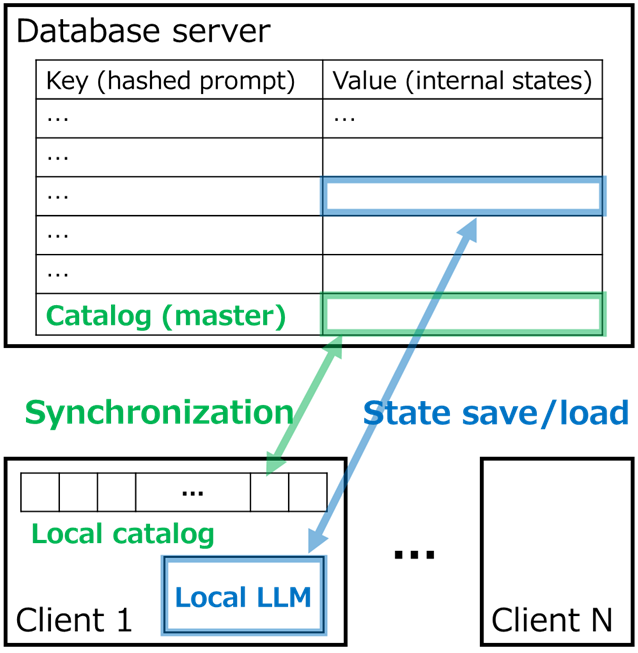}
  \caption{Data structures. The internal states and the master {\it catalog} are stored in the remote server. Local inference on the clients utilizes the cached states when available. Each local {\it catalog} is synchronized with the master on the server.}
  \label{fig:datastruct}
\end{figure}

To accelerate local LLM inference on resource-constrained edge devices, in this paper, internal states generated during prompt decoding (also known as prefill) are stored in a remote database server and shared among the same or different devices, as demonstrated in Figure \ref{fig:system}.
The underlying data structures are detailed in Figure \ref{fig:datastruct}.
By reusing these cached internal states for local LLM inference, the prompt decoding process for all or part of a given prompt can be bypassed.
This approach is beneficial if there is a sufficient similarity between the current and cached prompts, particularly when repeatedly utilizing the identical prompt templates or system instructions. 
This reduction in prompt decoding time is crucial, as Time to First Token (TTFT) directly impacts the user experience.

\subsection{Catalog}
The internal states, including the KV cache associated with a specific prompt, are hereafter referred to as the prompt cache.
The size of each prompt cache entry depends on the context length and the model architecture used; it is often significant (e.g., several megabytes or more) for resource-constrained devices.
This data volume imposes substantial communication overhead, specifically the latency associated with uploading and downloading cache entries, potentially outweighing the benefits of prompt cache sharing.
Since edge devices are often battery-powered and rely on wireless connectivity, remote database access must be minimized to reduce both power consumption and latency.

\begin{figure}[t]
  \centering
  \includegraphics[height=52mm]{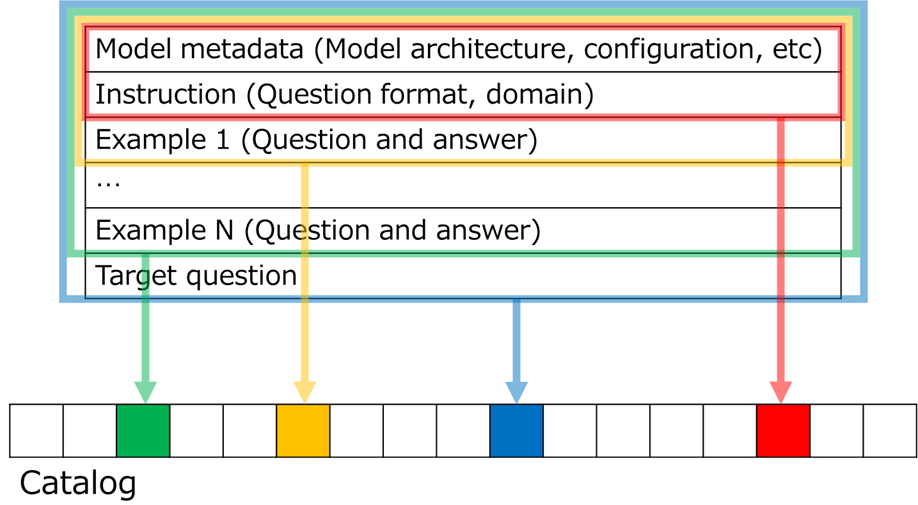}
  \caption{Example of {\it catalog}. Four distinct ranges of a prompt consisting of the instruction, few-shot examples, and target question are registered in the {\it catalog}.}
  \label{fig:prefix}
\end{figure}

To suppress unnecessary database access from edge devices, this paper introduces a Bloom-filter-based {\it catalog} that compactly summarizes the set of cached internal states stored on the remote server.
As shown in Figure \ref{fig:datastruct}, each client maintains its local {\it catalog}, which is synchronized with the master {\it catalog} on the server.
Figure \ref{fig:prefix} illustrates an example of the {\it catalog}, where each cell in the bottom part indicates whether the internal states corresponding to a given prompt are cached in the database server.

With the proposed {\it catalog}, local LLM inference on edge devices follows the steps below:
\begin{itemize}
\item Step 1: The edge device tokenizes the input prompt.
\item Step 2: It queries the local {\it catalog} to determine whether the remote server has cached the corresponding states for this prompt.
\item Step 3: If the local {\it catalog} hits, the edge device downloads the prompt cache from the server (see the blue arrow in Figure \ref{fig:datastruct}); otherwise, it decodes the prompt locally, uploads the resulting internal states to the server, and correspondingly updates its local {\it catalog} to reflect the new entry.
\item Step 4: It decodes response tokens and outputs them.
\end{itemize}
The local {\it catalog} is synchronized with the server asynchronously to reflect updates from other edge devices so as not to impact inference latency (see the green arrow in Figure \ref{fig:datastruct}).

To query or update the local {\it catalog} in Steps 2 and 3, a sequence of token IDs generated from a given prompt is hashed to generate a unique lookup key.
To ensure the integrity of the retrieved cache, additional metadata, such as the model name and its configuration parameters, is incorporated into the hash input, as shown in the top part of Figure \ref{fig:prefix}.
This approach distinguishes cached states from those generated under different model architectures or quantization settings.

\subsection{Partial Matching}
To further increase the cache hit ratio, the proposed {\it catalog} supports partial prompt matching.
This enables the reuse of cached prefix states by leveraging the logical structure of prompts.
Taking multiple-choice questions as an example, prompts typically consist of the following three parts:
\begin{itemize}
\item Instruction: Specifies the task format (i.e., multiple-choice question) and the domain (e.g., {\it astronomy}).
\item Few-shot examples: Contain $N$ pairs of question and answer.
\item Target question: The actual question to be answered.
\end{itemize}
In this case, multiple prompt ranges derived from a single input, such as the instruction alone, the instruction with examples, or the entire prompt, can be registered in the {\it catalog} during Step 3.
As shown in Figure \ref{fig:prefix}, we consider four distinct prompt ranges: 1) the instruction alone, 2) the instruction with the first example, 3) the instruction with all examples, and 4) the entire prompt.

Since a longer sequence of reused tokens results in a more substantial reduction in decoding time, the lookup strategy in Step 2 is adapted to examine these multiple ranges.
Specifically, if a match of sufficient length is identified among the examined ranges, the edge device initiates the retrieval of the longest matching prompt cache from the server.

\subsection{False Positives}
There is a possibility of false positives inherent in the Bloom-filter-based {\it catalog}. 
In such cases, edge devices may retrieve an unintended prompt cache from the server; however, since this cache will not match the current prompt, the decoding process cannot be bypassed and is instead fully computed at the edge.
Thus, these false positives do not impact the logical correctness while increasing the processing time.
This impact is discussed in Section \ref{sec:eval}.

\section{Implementation}\label{sec:imple}
Table \ref{tab:spec} details the specifications of the server and client devices.
The database server runs on a Raspberry Pi 5 Model B with 16GB DRAM, utilizing Redis 8.0.2 (see Cache Server in Figure \ref{fig:system}).
Redis snapshotting is disabled to minimize disk I/O overhead.
For the clients, local LLMs are deployed on two edge devices: Raspberry Pi Zero 2W with 512MB DRAM (low-end setting; see Clients in Figure \ref{fig:system}) and Raspberry Pi 5 Model B with 4GB DRAM (high-end setting).

A lightweight custom client program is implemented in C++ and compiled with g++ 14.2.0.
The LLM inference, Redis access, and Bloom filter functionalities are implemented using llama.cpp b7957 \footnote{\url{https://github.com/ggml-org/llama.cpp}}, Hiredis 1.2.0 \footnote{\url{https://github.com/redis/hiredis}}, and libbloom 2.0 \footnote{\url{https://github.com/jvirkki/libbloom}}, respectively.
Specifically, \texttt{llama\_state\_\allowbreak get\_data()} is used to extract internal states of local LLMs after prompt decoding, while \texttt{llama\_state\_set\_data()} is used to restore the saved states.
The Bloom filter is configured with a capacity of 1M entries and a target false-positive ratio of 1\%; in this setting, its size is only 1.20MB.

Finally, the server and clients are connected via 2.4GHz Wi-Fi 4, as shown in Figure \ref{fig:system}.

\begin{table}[t]
    \centering
    \caption{Specifications of server and client devices.}
    \label{tab:spec}
    \begin{tabular}{l|l|l|l}
    \hline\hline
    & Server & Client (high-end) & Client (low-end)\\
    \hline
    Platform & \multicolumn{2}{c|}{Raspberry Pi 5 Model B} & Zero 2W\\
    CPU & \multicolumn{2}{c|}{ARM Cortex-A76 2.4GHz} & Cortex-A53 1GHz\\
    DRAM & 16GB & 4GB & 512MB\\
    OS & \multicolumn{3}{c}{Raspberry Pi OS 13 (Trixie), 64-bit}\\
    \hline
    \end{tabular}
\end{table}

\section{Evaluations}\label{sec:eval}
\subsection{Evaluation Methodology}
The proposed distributed prompt caching is evaluated in terms of Time to First Token (TTFT) and Time to Last Token (TTLT) using two edge computing platforms: Raspberry Pi Zero 2W \footnote{\url{https://www.raspberrypi.com}} with 512MB DRAM (low-end) and Raspberry Pi 5 Model B with 4GB DRAM (high-end).
For the model configurations, we employ Gemma-3 270M \footnote{\url{https://deepmind.google/models/gemma/gemma-3}} for the low-end setting and the 1B variant for the high-end setting.
Greedy sampling is used as the decoding strategy across all experiments.

In the experiments, prompts are generated using the MMLU dataset, which consists of 57 domains \cite{Hendrycks21}.
Within each domain, the instruction and few-shot examples are shared across prompts. 
These examples and the target questions are sampled from the \texttt{val} and \texttt{test} sets of the same domain. 
Since the length of the question-answer pairs significantly affects processing time, our experiments specifically focus on pairs with 256 words or fewer; as a result, 6,434 prompts are tested in total.

Given a prompt, all or part of the prompt may hit the distributed prompt cache.
We consider the following five cases:
\begin{itemize}
\item Case 1: No cache hit (i.e., miss).
\item Case 2: Hits the instruction part only.
\item Case 3: Hits the instruction part and the first example.
\item Case 4: Hits the instruction part and all examples ($N=5$).
\item Case 5: Hits the entire prompt (i.e., full hit).
\end{itemize}
Cases 2, 3, 4, and 5 correspond to the red, yellow, green, and blue parts in Figure \ref{fig:prefix}, respectively.
To avoid excessive processing time, $N$ is set to 1 for the low-end setting, whereas it is set to 5 for the high-end setting, unless otherwise noted. 

\subsection{Evaluation Results}
\subsubsection{Benefit vs. Overhead}
\begin{figure}[t]
  \centering
  \includegraphics[height=56mm]{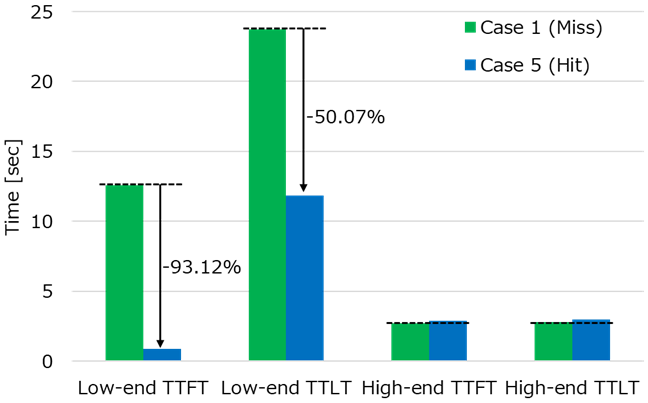}
  \caption{Performance comparison between cache miss and hit cases (based on Table \ref{tab:ttft1}).}
  \label{fig:ttft}
\end{figure}

Table \ref{tab:ttft1} presents the TTFT and TTLT for both low-end and high-end edge settings under Case 1 (cache miss) and Case 5 (full hit).
These latency values are averaged over 6,434 prompts.
Note that TTFT and TTLT include database access overhead over Wi-Fi (e.g., cache check, download, upload); a detailed breakdown of these components is provided later in this subsection.
In the low-end setting, the latency reduction is significant under Case 5; as shown in Figure \ref{fig:ttft}, TTFT and TTLT are reduced by 93.12\% and 50.07\% on average, respectively.
In contrast, the high-end setting shows a different trend; specifically, TTFT and TTLT increase by 7.08\% and 7.10\% on average, respectively, due to the database access overhead analyzed below.

\begin{table}[t]
    \centering
    \caption{TTFT and TTLT [sec] on low-end and high-end settings under Case 1 (cache miss) and Case 5 (full hit).}
    \label{tab:ttft1}
    \begin{tabular}{l|rrr|rrr}
    \hline\hline
    & \multicolumn{3}{c|}{TTFT} & \multicolumn{3}{c}{TTLT}\\
    Case & 1 & 5 & [\%] & 1 & 5 & [\%]\\
    \hline
    Low-end & 12.59 & 0.87 & 6.88 & 23.74 & 11.86 & 49.93\\
    High-end & 2.70	& 2.89 & 107.08 & 2.77 & 2.97 & 107.10\\
    \hline
    \end{tabular}
\end{table}

\begin{table*}[t]
    \centering
    \caption{Latency breakdowns [msec] on low-end and high-end settings under Case 1 (cache miss) and Case 5 (full hit).}
    \label{tab:ttft2}
    \begin{tabular}{l|rrrrrr|rrr}
    \hline\hline
    & Token & Bloom & P-decode & Redis & R-decode & Sample & $N$ & \# tokens & State size [MB]\\
    \hline
    Low-end (Case 1)  & 3.46 & 0.30 & 12580.85 & 2.42$\dagger$ & 11061.04 & 95.69 & 1 & 65.27 & 2.25\\
    Low-end (Case 5)  & 3.46 & 0.19 & 0.00 & 861.92 & 10904.67 & 84.82 & 1 & 65.27 & 2.25\\
    High-end (Case 1) & 1.61 & 0.00 & 2688.17 & 7.84$\dagger$ & 72.59 & 1.45 & 5 & 334.11 & 9.94\\
    High-end (Case 5) & 1.56 & 0.00 & 0.00 & 2887.04 & 78.12 & 1.67 & 5 & 334.11 & 9.94\\
    \hline
    \multicolumn{9}{l}{$\dagger$ Induced by rate false positives.}\\
    \end{tabular}
\end{table*}

Table \ref{tab:ttft2} details the latency breakdown for the low-end and high-end edge settings by considering the following six components:
\begin{itemize}
\item Token: The time required to tokenize a given prompt.
\item Bloom: The time required to determine whether the remote server has cached a given prompt by querying the local {\it catalog}.
\item P-decode: The time required to decode a given prompt.
\item Redis: The time required to download/upload a corresponding prompt cache from/to the remote database server over Wi-Fi.
\item R-decode: The time required to decode a response token. In the case of multiple-choice questions, the output token length is typically one (e.g., ``A’’, ``B’’, ``C’’, ``D’’).
\item Sample: The time required to sample the response token using a greedy sampling strategy.
\end{itemize}
Note that the TTFT under Case 1 comprises Token, Bloom, and P-decode, whereas under Case 5 it includes Token, Bloom, and Redis.
The TTLT consists of R-decode and Sample in addition to the components comprising the TTFT.
The Redis access time depends on the cache entry size and available network bandwidth, where the cache entry size is determined by the model used.
In this experiment, the cache entry size is 2.25MB for the 270M model (low-end setting), whereas it is 9.94MB for the 1B model (high-end setting).

As shown in Table \ref{tab:ttft2}, for the low-end setting, P-decode requires 12.58 seconds under Case 1, which is replaced by Redis access requiring only 0.86 seconds under Case 5.
Consequently, the proposed distributed prompt caching significantly reduces both TTFT and TTLT for the low-end setting in Case 5.
In contrast, for the high-end setting, P-decode requires 2.69 seconds under Case 1, whereas Redis access requires 2.89 seconds under Case 5.
This indicates that the benefit of a prompt cache hit is offset by the Redis access overhead over Wi-Fi in the high-end setting.
In summary, the distributed prompt cache is highly effective for local LLM inference on resource-constrained computing platforms, such as Raspberry Pi Zero 2W.

\subsubsection{Benefit of Partial Matching}
Here, we discuss how partial matching of the prompt cache affects the total decoding time (i.e., the sum of P-decode and R-decode).
For this analysis, as shown in Figure \ref{fig:prefix}, we selected a single prompt comprising an instruction, five examples (i.e., $N=5$), and a target question from the {\it astronomy} domain.
The total number of tokens in this prompt is 405.
In this analysis, the numbers of matched tokens are 1, 10, 57, 340, and 405 for Cases 1, 2, 3, 4, and 5, respectively.

\begin{figure}[t]
  \centering
  \includegraphics[height=63mm]{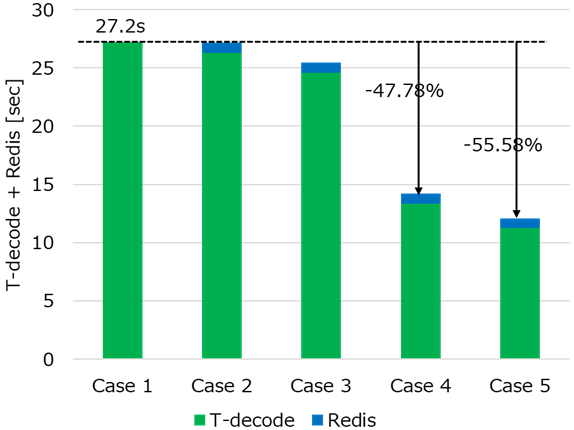}
  \caption{Performance comparison between partial matching cases on low-end setting (based on Table \ref{tab:prefix}).}
  \label{fig:breakeven}
\end{figure}

Table \ref{tab:prefix} presents the total decoding time (excluding Redis access overhead) under these five cases for the low-end and high-end settings.
As the number of matched tokens increases, the total decoding time decreases as expected. 
For the low-end setting, Cases 2, 3, 4, and 5 reduce the total decoding time by 0.92, 2.61, 13.86, and 15.98 seconds, respectively, compared to Case 1.
Figure \ref{fig:breakeven} shows the performance comparison on the low-end setting assuming that the Redis access overhead is 0.86 seconds (represented by the blue part in the graph) as mentioned earlier.
Under this condition, the partial prompt caching is proven to be effective especially for Cases 4 and 5 even when this overhead is taken into account.

\begin{table}[t]
    \centering
    \caption{Total decoding time [msec] on low-end and high-end settings under partial matching cases.}
    \label{tab:prefix}
    \begin{tabular}{l|rrrr}
    \hline\hline
    & \# matched & \% matched & T-decode \\
    \hline
    Low-end (Case 1) & 1 & 0.25 & 27203.96\\
    Low-end (Case 2) & 10 & 2.47 & 26288.23\\
    Low-end (Case 3) & 57 & 14.07 & 24590.09\\
    Low-end (Case 4) & 340 & 83.95 & 13344.96\\
    Low-end (Case 5) & 405 & 100.00 & 11220.95\\
    High-end (Case 1) & 1 & 0.25 & 3361.88\\
    High-end (Case 2) & 10 & 2.47 & 3280.38\\
    High-end (Case 3) & 57 & 14.07 & 2918.08\\
    High-end (Case 4) & 340 & 83.95 & 643.35\\
    High-end (Case 5) & 405 & 100.00 & 62.9\\
    \hline
    \end{tabular}
\end{table}

\subsubsection{Benefit of Local Bloom Filter}
In the distributed prompt caching mechanism, clients first query the local {\it catalog} to determine whether the remote server has cached a given prompt, thereby suppressing unnecessary database access.
As mentioned above, the Redis access overhead is not negligible (e.g., 0.86 seconds for the low-end setting).
Without the proposed {\it catalog}, every LLM inference would incur this overhead, which would offset the benefit of the distributed prompt caching, especially under a low cache-hit scenario.
By introducing the {\it catalog}, the Redis access overhead is incurred only when a corresponding prompt cache is identified as available, regardless of the actual hit ratio, with the exception of Bloom filter false positives, which are discussed in the next subsection.

\subsubsection{Impact of Bloom Filter False Positives}
The {\it catalog} is implemented based on the assumption that the Bloom filter false positive ratio is 1\%.
In the event of a false positive, the TTFT and TTLT in Case 1 would incur the Redis access overhead.
Specifically, for the low-end setting, the expected TTFT and TTLT for Case 1 increase by only 0.86 $\times$ 0.01 seconds, which has a negligible impact on the break-even point of the proposed distributed prompt cache (see also Figure \ref{fig:breakeven}).

\subsection{Practical Considerations}
In Figure \ref{fig:system}, local LLMs are deployed on two Raspberry Pi Zero 2W boards (the left and right nodes).
As demonstrated above, the proposed distributed prompt caching is particularly beneficial for these devices due to their severe resource constraints.
Specifically, the limited CPU performance, reflected in long TTFT and TTLT, is mitigated by reusing internal states from the same or different nodes.
Furthermore, the limited DRAM capacity (e.g., only 512MB) is addressed by utilizing the storage of a remote server.
We employ an off-the-shelf Redis running on Raspberry Pi 5 (the middle node) as a {\it cache box}.
As shown in Figure \ref{fig:system}, by simply adding this {\it cache box} to the setup, overall performance is significantly improved.
Please note that local LLM inference on the left and right nodes remains functional even if the middle node is unavailable, although performance will be significantly degraded due to the loss of the proposed distributed caching.

\section{Summary}\label{sec:conc}
Since local LLM inference on low-end edge devices imposes a severe performance bottleneck, this paper proposes a distributed prompt caching mechanism by extending the concept of prompt prefix caching to a multi-device setting to fully utilize prompt similarity.
While our approach introduces communication overhead associated with state sharing over Wi-Fi, the experiments demonstrate that it is highly effective in the low-end setting, where TTFT and TTLT are reduced by 93.12\% and 50.07\% on average, respectively.
In contrast, the approach is less beneficial for high-end edge devices. 
Furthermore, the experiments show that partial matching of our distributed cache effectively increases cache reuse opportunities while reducing latency.
Finally, thanks to the proposed local {\it catalog}, communication takes place only when a corresponding cache is likely to exist on the server.
This suppresses unnecessary communication overhead, favorably shifting the break-even point of this approach.

A demonstration video of the proposed system running on Raspberry Pi Zero 2W is available at: \url{https://www.youtube.com/watch?v=3qcExKXdkvQ}


\bibliographystyle{unsrt}  
\bibliography{references}

\end{document}